\documentclass[conference]{IEEEtran}
\IEEEoverridecommandlockouts
\usepackage{cite}
\usepackage{enumitem}
\usepackage{multirow}
\usepackage{float}
\usepackage{amsmath,amssymb,amsfonts}
\usepackage{algorithmic}
\usepackage{graphicx}
\usepackage{textcomp}
\usepackage{xcolor}
\usepackage{subfigure}
\usepackage{placeins}
\usepackage{stfloats}
\usepackage{mathtools}
\def\BibTeX{{\rm B\kern-.05em{\sc i\kern-.025em b}\kern-.08em
    T\kern-.1667em\lower.7ex\hbox{E}\kern-.125emX}}
\begin{document}

\title{MUSEFood: Multi-sensor-based Food Volume Estimation on Smartphones\\}

% \author{\IEEEauthorblockN{1\textsuperscript{st} Junyi Gao}
% \IEEEauthorblockA{\textit{Key Lab of High Confidence Software Technologies} \\
% \textit{Peking university}\\
% Beijing, China \\
% buaagjy@buaa.edu.cn}
% \and
% \IEEEauthorblockN{2\textsuperscript{nd} Weihao Tan}
% \IEEEauthorblockA{\textit{Key Lab of High Confidence Software Technologies} \\
% \textit{Peking university}\\
% Beijing, China \\
% email address}
% \and
% \IEEEauthorblockN{3\textsuperscript{rd} Liantao Ma}
% \IEEEauthorblockA{\textit{dept. name of organization (of Aff.)} \\
% \textit{name of organization (of Aff.)}\\
% City, Country \\
% email address}
% \and
% \IEEEauthorblockN{4\textsuperscript{th} Yasha Wang}
% \IEEEauthorblockA{\textit{dept. name of organization (of Aff.)} \\
% \textit{name of organization (of Aff.)}\\
% City, Country \\
% email address}
% \and
% \IEEEauthorblockN{5\textsuperscript{th} Wen Tang}
% \IEEEauthorblockA{\textit{dept. name of organization (of Aff.)} \\
% \textit{name of organization (of Aff.)}\\
% City, Country \\
% email address}
% }

\author{\IEEEauthorblockN{Junyi Gao\IEEEauthorrefmark{2}\IEEEauthorrefmark{4}\IEEEauthorrefmark{6}\IEEEauthorrefmark{7}, Weihao Tan\IEEEauthorrefmark{2}\IEEEauthorrefmark{6}\IEEEauthorrefmark{7}, Liantao Ma\IEEEauthorrefmark{2}\IEEEauthorrefmark{6}, Yasha Wang\IEEEauthorrefmark{2}\IEEEauthorrefmark{3}\IEEEauthorrefmark{4}\IEEEauthorrefmark{1} and Wen Tang\IEEEauthorrefmark{5}}
\IEEEauthorblockA{
\IEEEauthorrefmark{2}Key Laboratory of High Confidence Software Technologies, Ministry of Education, Beijing 100871, China\\
\IEEEauthorrefmark{3}National Engineering Research Center for Software Engineering, Peking University, Beijing 100871, China\\
\IEEEauthorrefmark{4}Peking University Information Technology Institute (Tianjin Binhai), Tianjin 300450, China\\
\IEEEauthorrefmark{6}School of Electronics Engineering and Computer Science, Peking University, Beijing 100871, China\\
\IEEEauthorrefmark{5}Department of Nephrology, Peking University Third Hospital, Beijing 100191, China\\
\IEEEauthorrefmark{7}Same contribution\\
\IEEEauthorrefmark{1}Corresponding to wangyasha@pku.edu.cn}}

\maketitle

\begin{abstract}
Researches have shown that diet recording can help people increase awareness of food intake and improve nutrition management, and thereby maintain a healthier life. Recently, researchers have been working on smartphone-based diet recording methods and applications that help users accomplish two tasks: record what they eat and how much they eat. Although the former task has made great progress through adopting image recognition technology, it is still a challenge to estimate the volume of foods accurately and conveniently. In this paper, we propose a novel method, named \textit{MUSEFood}, for food volume estimation. \textit{MUSEFood} uses the camera to capture photos of the food, but unlike existing volume measurement methods, \textit{MUSEFood} requires neither training images with volume information nor placing a reference object of known size while taking photos. In addition, considering the impact of different containers on the contour shape of foods, \textit{MUSEFood} uses a multi-task learning framework to improve the accuracy of food segmentation, and uses a differential model applicable for various containers to further reduce the negative impact of container differences on volume estimation accuracy. Furthermore, \textit{MUSEFood} uses the microphone and the speaker to accurately measure the vertical distance from the camera to the food in a noisy environment, thus scaling the size of food in the image to its actual size. The experiments on real foods indicate that \textit{MUSEFood} outperforms state-of-the-art approaches, and highly improves the speed of food volume estimation.
\end{abstract}

\begin{IEEEkeywords}
food volume estimation, diet management, image segmentation, smartphone sensing
\end{IEEEkeywords}

\section{Introduction}
People's food intake has been proven to have a major impact on health. According to medical researches, many chronic diseases such as diabetes and kidney disease have been attributed to dietary factors\cite{who2003diet}. Diet recording can help people to maintain a healthy diet and thus improve their health status\cite{ortega1998difference}.

With the popularity of smartphones, a number of apps for diet recording (e.g. MyFitnessPal\footnote{website: https://www.myfitnesspal.com/}, LoseIt\footnote{website: https://www.loseit.com/}, etc) have been developed to help users assess their food intake. A number of studies have demonstrated that smartphone-based diet record APPs can significantly increase the user's awareness of food intake, improve their nutritional management, and thereby improve their health\cite{coughlin2015smartphone}.

Smartphone-based diet recording methods and applications help users accomplish two tasks: record what they eat and how much they eat. For the former task, researchers have proposed a series of methods based on image recognition technology in recent years\cite{chen2016deep, ciocca2017food}, which can accurately identify the food categories and greatly simplify the manual workload of users. However, for the latter task, existing APPs do not provide much support. Users still have to estimate foods' weight or volume according to their own personal experience and then input data into the APPs manually. Studies have shown that the average of the error in the manual estimation is above 20\%\cite{godwin2004accuracy, hernandez2006portion}, leading to the estimation of food volume becoming the bottleneck of smartphone-based dietary recording methods. To solve this problem, researchers have started developing affordable smartphone aided food volume estimation solutions in recent years\cite{akpa2017smartphoneF, liang2017deep, dehais2017two, chung2017glasses}. However, existing works are limited in terms of user convenience and accuracy, mainly because they cannot effectively address the following two major challenges.

\begin{itemize}[leftmargin=*]
\item \textbf{$C_{1}$: Converting the relative size of the food in an image to its actual size conveniently and accurately.} From the image, we can see the relative size of the food comparing to other items such as containers, chopsticks, forks, etc., which are also present in the same image. However, since the actual physical size corresponding to each pixel is unknown, the actual volume of the food cannot be obtained from the image alone. In order to solve this problem, some existing works require the user to place some reference objects of known size (e.g. credit card\cite{dehais2017two}, chopsticks\cite{akpa2017smartphoneF} or fingers\cite{pouladzadeh2015foodd}) together with the food into the same photo. However, these methods require the user to always carry the reference objects, which causes inconvenience. Some other methods use images with volume information (e.g. depth images taken by special device\cite{meyers2015im2calories} or up to 16 images of the same food taken from different heights and angles\cite{allegra2017multimedia}) to train their models. These methods do not require the placement of reference objects, however, they can only be used for the same type of food as the food in the training set. And because of the high cost of expanding the training set, the scope of their application is quite limited.
\item \textbf{$C_{2}$: Segmenting the food in the image from the background accurately and efficiently.} The segmentation of food from the background is an important basis and prerequisite for constructing a food volume model. The accuracy and efficiency of segmentation has a critical impact on the overall accuracy and efficiency of food volume estimation. Recent researchers combine \textit{Convolutional Neural Networks}(CNN) and \textit{Grabcut}\cite{okamoto2016automatic,liang2017deep} or use point matching\cite{dehais2017two} to segmented food items. In order to achieve better segmentation accuracy, these methods have strict requirements on image capture methods and image quality (e.g. good camera exposure and focus, strong, variable texture in the scene and limited secular reflections and even specified shooting angles), which pose significant inconvenience and robustness issues in actual use\cite{dehais2017two}.
\end{itemize}

\begin{figure}[htb]
  \centering
  \subfigure[The foods contained in round bowls are usually rounded at the edges]{
      \includegraphics[scale=0.25]{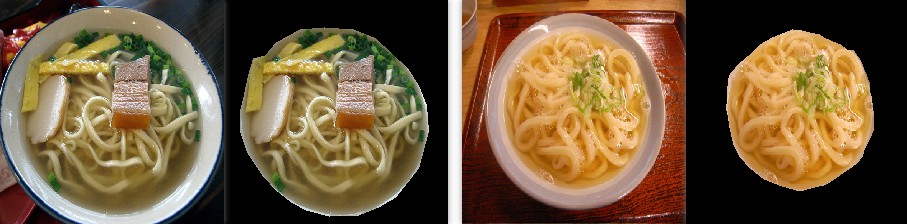}
      \label{fig:bowls}
  }
  \subfigure[The shapes of edges of the foods contained in plates may vary greatly]{
      \includegraphics[scale=0.25]{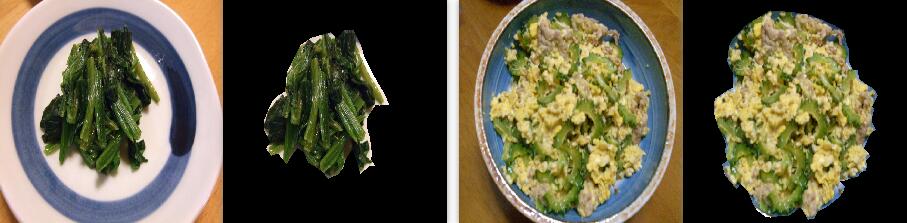}
      \label{fig:plates}
  }
  \caption{The shapes of food contours are often related with the shapes of containers}
  \label{fig:plate and bowl}
\end{figure}

With the objective of taking full consideration of the above mentioned challenges, we have the following observations: 1). The existing methods mainly use cameras. However, other sensors are generally installed on smartphones, and the use of multiple sensors may improve the accuracy of the food volume estimation. 2). Compared with the conventional computer vision-based methods, using deep learning-based methods to directly segment food items will greatly improve the accuracy and speed of segmentation. Furthermore, we notice that the shapes of food contours are often related with the shapes of containers. Figure \ref{fig:plate and bowl} shows an example. By utilizing the relationship between the shapes of food contours and the categories of the food containers, the accuracy of food segmentation can be improved.

Based on the above observations, we propose a \textbf{\underline{Mu}}lti-\textbf{\underline{Se}}nsor-based \textbf{\underline{Food}} volume estimation method on smartphones, \textit{MUSEFood}. \textit{MUSEFood} collects data from multiple sensors on smartphones, processes and calculates sensing data, and aggregates the processing results to calculate the actual food volume. Our main contributions are summarized as follows:

\begin{itemize}[leftmargin=*]
\item We propose a method for food volume estimation, \textit{MUSEFood}, which involves three stages: sensing, data processing and data aggregation. To the best of our knowledge, our method is the first to require neither training images with volume information nor setting a reference object of known size while taking photos of foods.
\item We introduce and further extend two technical mechanisms to calculate food volume more accurately, efficiently and versatilely. Specifically, 1).We utilize echo ranging method to accurately measure the vertical distance between the camera and the food by using the speaker to emit a \textit{Maximum Length Sequence} sound wave and using the microphone to sense the reflected wave. Based on this distance, we accurately calculate the actual volume of food without using an object of known size as a reference (i.e., addressing $C_{1}$). 2).We adopt a multi-task learning framework based on \textit{Fully Convolutional Networks} to segement food items. The proposed deep learning framework makes full use of the relationship between the shapes of food contours and the shapes of containers, resulting in accurate and efficient segmentation. This framework helps to achieve more accurate food volume modeling (i.e., addressing $C_{2}$).
\item We use real foods to verify and evaluate \textit{MUSEFood}. Compared with the state-of-the-art models, \textit{MUSEFood} is more accurate and faster in estimating the food volume with no need of using reference objects. Besides, \textit{MUSEFood} is applicable for both bowls and plates and also for different foods with irregular shapes, which is universal for more situations.
\item Along with this paper, we release the food dataset, SUEC Food\footnote{SUEC Food dataset and the source code in this paper are available at https://github.com/MUSEFood/MUSEFood}, which contains 600 segmented food images annotated manually and 31395 images annotated using \textit{Grabcut} method. All the images are from the UEC Food-256 dataset\cite{kawano2014automatic}. The foods in SUEC Food are mainly Asian, and the containers of foods are varied(e.g. plates, bowls and cups). Prior to this work, there is only one public food image dataset with segmentation labels, which only contains western food(e.g. pizzas and burgers) served on plates\cite{ciocca2017food}.
\end{itemize}

\section{Related Work}
There have been several recent attempts to automatically estimate food volume using smartphones. To achieve this, the proposed systems have to involve three modules: food item segmentation, actual size scaling and food modeling .

\textit{Food Item Segmentation}: The objective of the food item segmentation step is to detect the exact location of the food items in the images. Researchers used to use traditional computer vision methods such as \textit{Grabcut}\cite{rother2004grabcut} and \textit{Hough Transform}\cite{duda1971use} to segment food items. Recently, researchers start to combine deep-based methods and conventional computer vision systems to solve this problem. Some researchers \cite{liang2017deep, okamoto2016automatic} use CNN to draw bounding boxes around food items, and use \textit{Grabcut} method to segment food items from bounding boxes. Another researchers\cite{dehais2017two, rahman2012food} use point matching methods to locate food items in images, which require two images taken from different angles to match similar objects, and then get segmented food items. However, these methods are not robust and convenient enough due to the requirement of shooting angle and the background of images. Most importantly, for both methods, the accuracy of the segmentation still needs to be improved, especially for foods with complex and irregular shapes. Besides, these algorithms require a large amount of computation, which will take a lot of time and affect the user experience in practice.

\textit{Actual Size Scaling}: The actual size scaling module is used to obtain the actual size of each pixel of the image in order to estimate the actual size of the food in the image. Many researchers choose to place a reference object of known size next to the food and compare the relative size of the food to the reference object in the image to calculate the actual volume of food. There are various reference objects used to estimate food volume, such as chessboard-like marker\cite{woo2010automatic}, credit card\cite{okamoto2016automatic}, chopsticks\cite{akpa2017smartphoneF}, finger\cite{almaghrabi2012novel} and so on. However, different kinds of reference objects bring different problems. For reference objects such as credit card, users need to carry them at any time, which is inconvenient in actual use. Though,  indeed,  plates,  chopsticks  and  fingers are reference objects that do not have to be carried by users. However, we can't guarantee that everyone uses the same length of chopsticks or have the same size of fingers, which may cause huge errors. Other researchers use training images with volume information (e.g. depth map\cite{meyers2015im2calories, allegra2017multimedia} or images of the same food taken from from different heights and angles\cite{allegra2017multimedia} ) to avoid using reference objects. These methods require special devices to collect food data and build their own dataset, and the foods must be placed on plates, so that they are greatly limited in application.

\textit{Food Modeling}: Food modeling is the module that builds a 3D model of food based on 2D images. The volume of food can be estimated using this 3D model. In some early food modeling methods\cite{almaghrabi2012novel}, food items are modeled as regular geometries. Since the shapes of foods are not always regular, these methods may cause huge errors. Some researches use deep neural networks to reconstruct depth maps of foods\cite{meyers2015im2calories}. However, these methods require special training images as mentioned above. Recently, some researchers use stereo matching based modeling methods to build the point cloud of foods\cite{dehais2017two,hassannejad2017new}. Other researchers build different food models based on different containers\cite{akpa2017smartphoneF}. These methods require users to take 2-3 photos for each food. In this paper, we mainly focus on the improvements for food item segmentation and actual size scaling. For food modeling, based on the previous researches of 3D reconstruction, we propose a simplified method to quickly build food models using two food images, which is applicable to different containers and different foods with irregular shapes.

% \subsection{Mobile-sensor-based Distance Measurement}

% As mentioned above, many systems only use the camera on the smartphone to take food pictures so that they have to use an object of known size as a reference to calculate the actual food size. Instead of using an object, the distance from the camera to the object can also be used as a reference. Though some researchers\cite{holzmann2012measuring} developed methods to use single camera to measure distance. These methods are only applicable for long distance measurement, and is hard to apply in food volume estimation. The \textit{AR Ruler} based on \textit{ARKit} in iOS developed by Apple can measure the length of an object in real time. However, instead of measuring the vertical distance, it can only measure the straight line distance of the object. The food of irregular size cannot be measured, which makes this method unsuitable for food volume estimation. Some researchers develope distance measurement methods using sound wave. Dunn C and Król D et.al\cite{dunn1993distortion, simon2004sensor} use \textit{Maximum Length Sequence} to measure distance. These methods make it possible to measure precise distance on smartphones.

\section{Methods}
\subsection{Overview}\label{AA}
In \textit{MUSEFood}, the whole task of food volume estimation is divided into three steps: Sensing, Data Processing and Data Aggregation. First, the user uses smartphone camera to take food photos. The speaker of the smartphone emits \textit{Maximum Length Sequence} of a specific length while the user is taking photos. Then the microphone receives the echo of the signal. In the second step, Data Processing, the food in the image is segmented from background. Meanwhile, we analyze the received echo signal to calculates the vertical distance from the smartphone lens to the food. In the third step, Data Aggregation, we combine the processed data from images and sound waves to build food model and finally estimate food volume.

\subsection{Sensing}
Different from other food volume estimation methods\cite{akpa2017smartphoneF, liang2017deep, dehais2017two, chung2017glasses} which only use the camera to collect data, \textit{MUSEFood} uses multiple sensors on smartphone including camera and microphone. This section is divided into two steps: Food Image Sensing and Audio Sensing.

\subsubsection{Food Image Sensing}
In order to accurately calculate the volume of the food, the user is required to take two food photos. First, the user needs to take a top-down photo of the food while ensuring that the phone and the surface of the food container are parallel to each other. The user then needs to take a side photo of the food while ensuring that the phone is perpendicular to the surface of the container.

\subsubsection{Audio Sensing}
We aim to develop a method that uses the distance from the smartphone to desktop as the reference instead of placing an object of known size. A non-invasive method for distance measurement is echo ranging. We use the Maximum Length Sequence (MLS) as the audio signal for ranging.

The MLS measurement technique is widely used in the field of acoustics for measuring concert hall acoustics\cite{ando2012concert} as well as loud-speaker transfer functions. The MLS measurement is particularly attractive because of the higher computational efficiency of processing and distortion and noise immunity compared with other audio signals such as chirp signal and sine signal\cite{dunn1993distortion}. 

\begin{figure}[ht]
\centering
\includegraphics[scale=0.11]{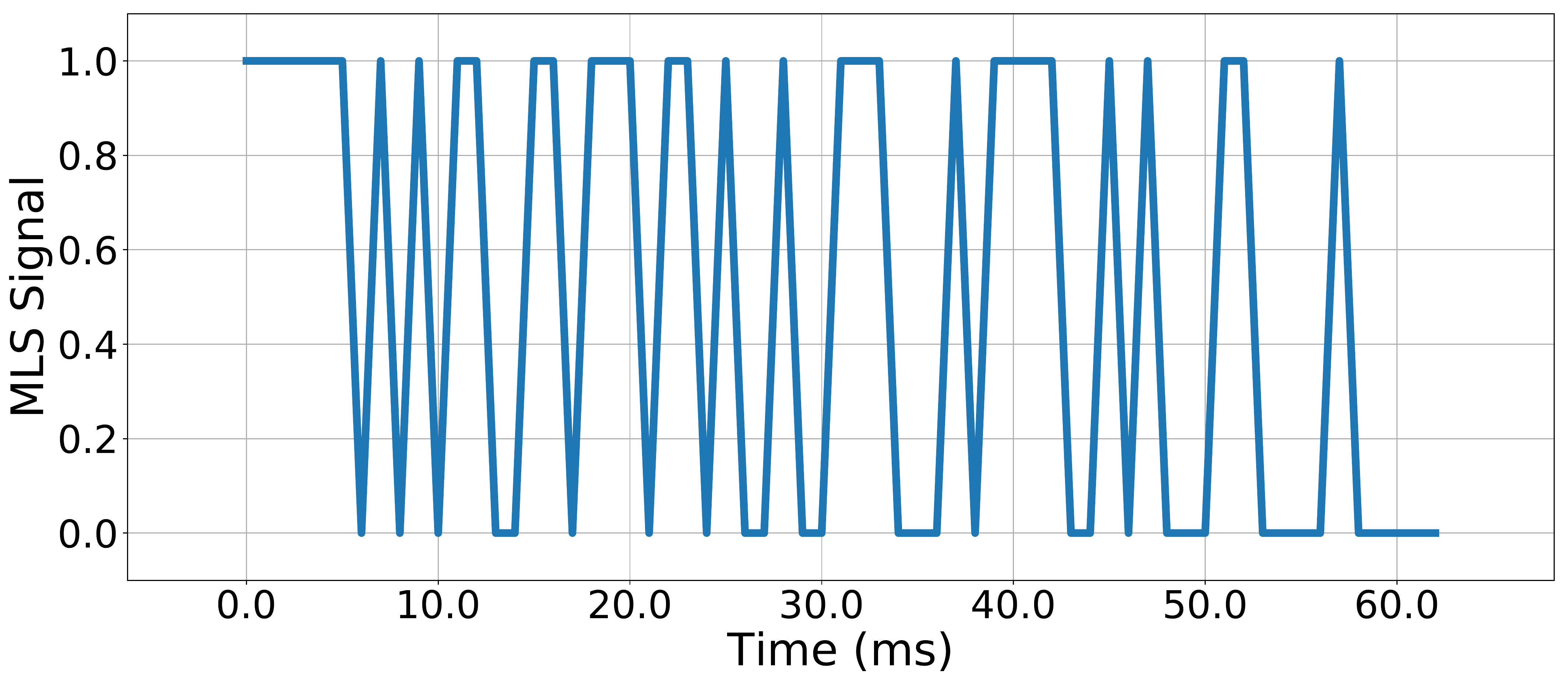}
\caption{MLS signal}
\label{fig:mls_signal}
\end{figure}

The MLS is a pseudorandom binary sequence (PRBS) composed of a sequence of 1 and 0 (Fig. \ref{fig:mls_signal}), generated recursively using a series of digital shift registers with selected XOR feedback. The length $L_{s}$ of the MLS given by $L_{s}=2^n-1$ with $n$ denoting the order of the sequence and also the number of digital shift registers.

While the user is taking a top-down photo, the microphone sensor is activated to start recording. Then the phone speaker emits a specific length $L_{s}$ of MLS audio signal $s_{o}$. When the user finishes photoing, the microphone stops recording. This recorded audio signal $s_{r}$ contains the original MLS signal played by the speaker and the audio signal reflected by obstacles (i.e. body, desktop and wall). These two signals will be used to calculate the vertical distance from the smartphone lens to the desktop while taking the photo.

\subsection{Data Processing}
In this section, we will process the collected sensing data and extract information from raw data to calculate the food volume. It is divided into two steps: processing the audio signal to obtain the vertical distance from the phone to the desktop, and segmenting the images to obtain segmented food items in the images.

\subsubsection{Audio Signal Processing For Distance Measurement}

\begin{figure}[ht]
\centering
\includegraphics[scale=0.13]{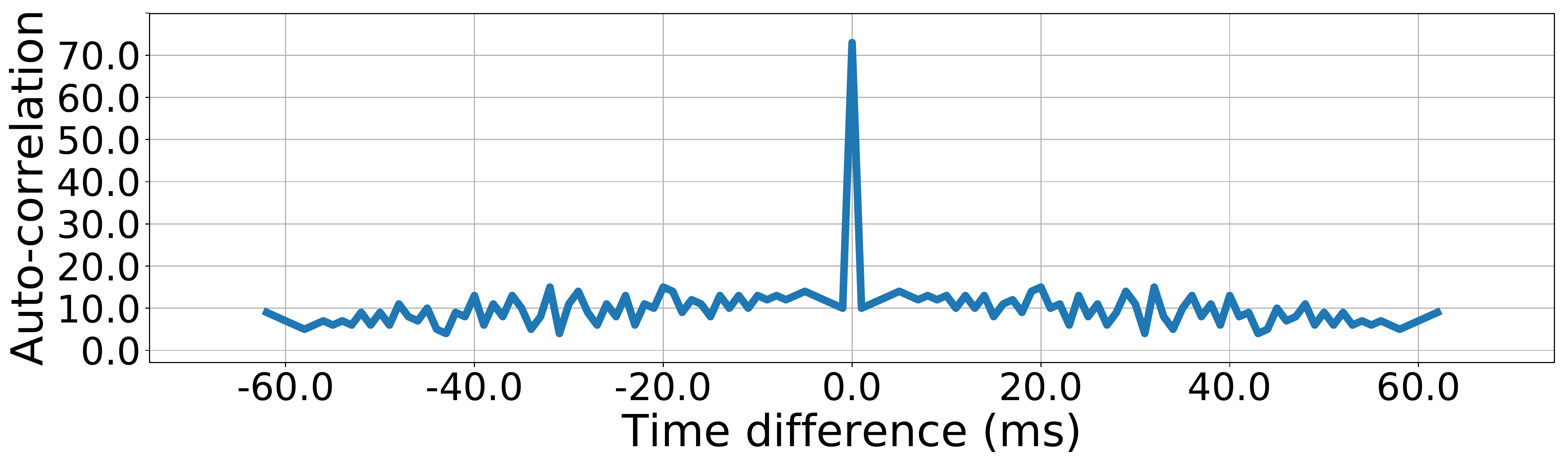}
\caption{Auto-correlation of MLS}
\label{fig:mls_auto}
\end{figure}

As mentioned above, MLS is essentially pseudo-random binary sequences which yield a unit impulse upon circular autocorrelation (Fig. \ref{fig:mls_auto}). Its periodic autocorrelation function $\phi_{nn}(l)$ is the two valued Kronecker function $\delta(l)$:
\begin{equation}
    \phi_{nn}(l) = \frac{L+1}{L}\delta(l)-\frac{1}{L}
\end{equation}
with
\begin{equation}
    \delta(l) = 
    \begin{cases}
        1,& \text{for }l=0\\
        0,& \text{for }l\neq0
    \end{cases}
\end{equation}
where $l$ is calculated modulo $L_{s}$. A longer MLS sequence produces a smaller error in the autocorrelation function.

The time when the microphone receives the MLS signals can be retrieved from the cross-correlation function between the recorded signal $s_{r}$ and the original signal $s_{o}$.

\begin{figure}[ht]
\centering
\includegraphics[scale=0.13]{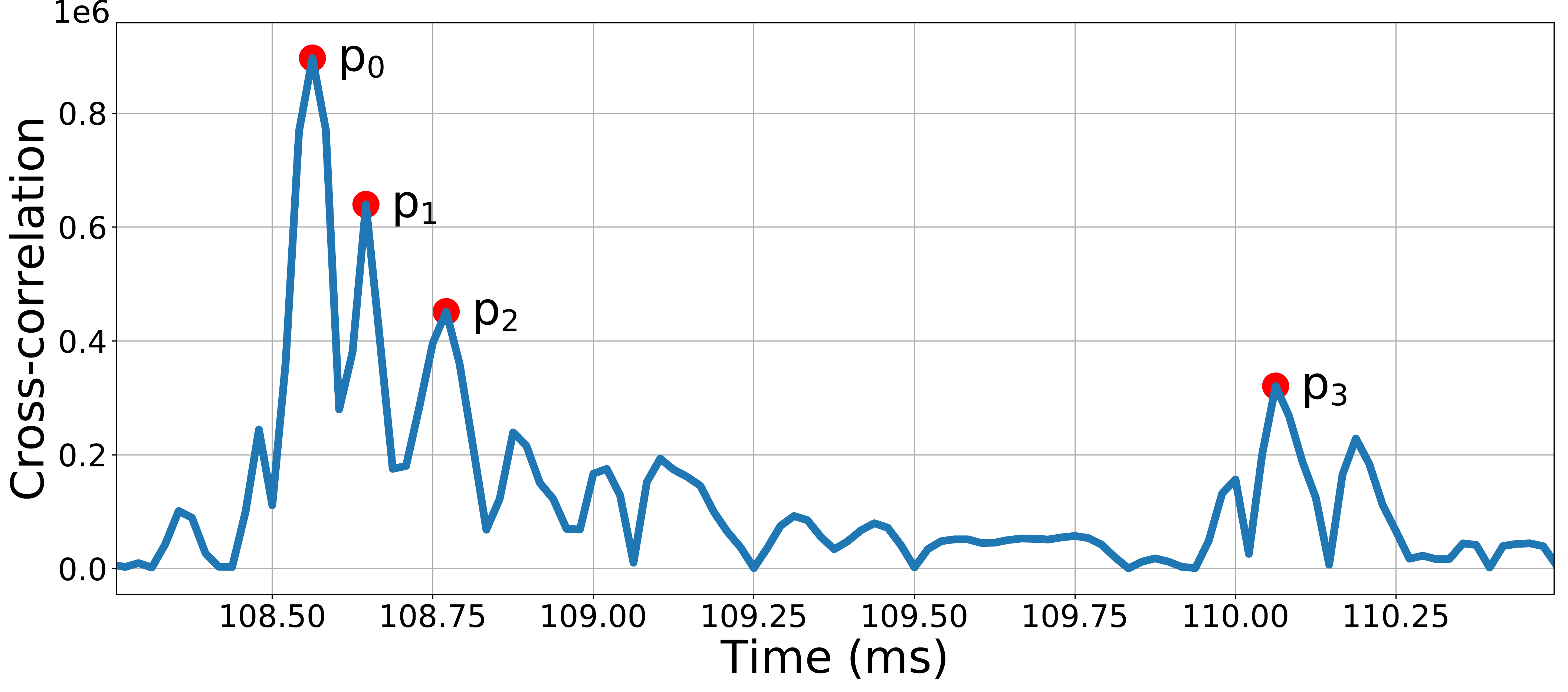}
\caption{Cross-correlation of the recorded signal and the original MLS signal}
\label{fig:mls_corr}
\end{figure}

Since the recorded audio signal $s_{r}$ contains the original MLS signal and the audio signal reflected by obstacles, there are multiple peaks in the cross-correlation results. The highest peak represents the recorded original MLS signal, and each of the remaining peaks represents an echo that is reflected by an obstacle. As shown in Figure \ref{fig:mls_corr}, $p_{0}$ represents the recorded original MLS signal, and $p_{3}$ represents the MLS signal reflected by desktop. $p_{1}$ and $p_{2}$ represents the signal reflected by other closer obstacles such as the user's body. $t_{0},t_{1},t_{2},t_{3}$ are the times corresponding to the peaks, which indicate the moments when the microphone recorded the reflected echos. 

\begin{figure}[ht]
\centering
\includegraphics[scale=0.15]{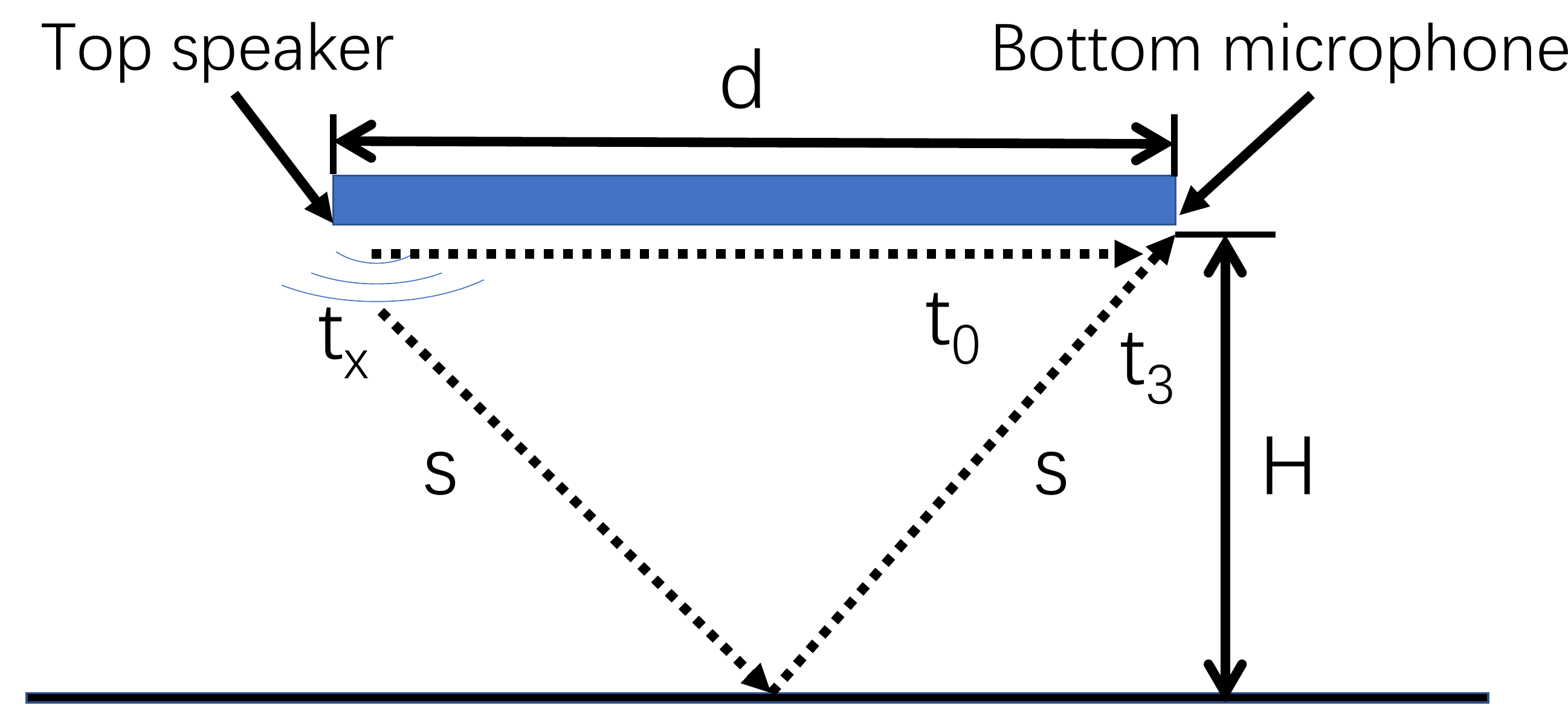}
\caption{Echo ranging on the smartphone using MLS}
\label{fig:dis_measure}
\end{figure}

As shown in Figure \ref{fig:dis_measure}, the vertical distance of the phone from the desktop while shooting can be caluculated as:
\begin{equation}
    \begin{cases}
        (t_{0}-t_{x})v &= d\\
        (t_{3}-t_{x})v &= 2s\\
        H^2 + (\frac{d}{2})^2 &= s^2
    \end{cases}
    \rightarrow
    H=\frac{\sqrt{(vt_{3}-vt_{0}+d)^2-d^2}}{2}
\end{equation}
where $t_{x}$ is the moment when the speaker emits the MLS signal, $2s$ is the distance traveled by echo $p_{3}$, $d$ is the distance from the phone speaker to the microphone, $v$ is the sonic speed and $H$ is the distance from the phone to the desktop to be measured.

\subsubsection{Food Item Segmentation}

Fully Convolutional Networks (FCN)\cite{long2015fully} is an end-to-end system with input as a image, and output as pixel-wise predictions for the image. However, to the best of our knowledge, it has not been used in food volume estimation so far. In this work, we use FCN for food image segmentation instead of the CNN and \textit{Grabcut} methods used in traditional food volume estimation methods\cite{liang2017deep, okamoto2016automatic}.

% \begin{figure}[htbp]
% \centering
% \subfigure[Original food image]
% {
% 	\includegraphics[width=0.33\linewidth, height=50pt]{food.jpg}
%     \label{subfig:food}
% }
% \subfigure[Segmented food image]
% {
% 	\includegraphics[width=0.33\linewidth, height=50pt]{food_segmentation.jpg}
% 	\label{subfig:food_segementation}
% }
% \caption{Food image segmentation}
% \label{fig:food_segment}
% \end{figure}
\begin{figure*}[htb]
  \centering
  \subfigure[MFCN-A]{
      \includegraphics[scale=0.17]{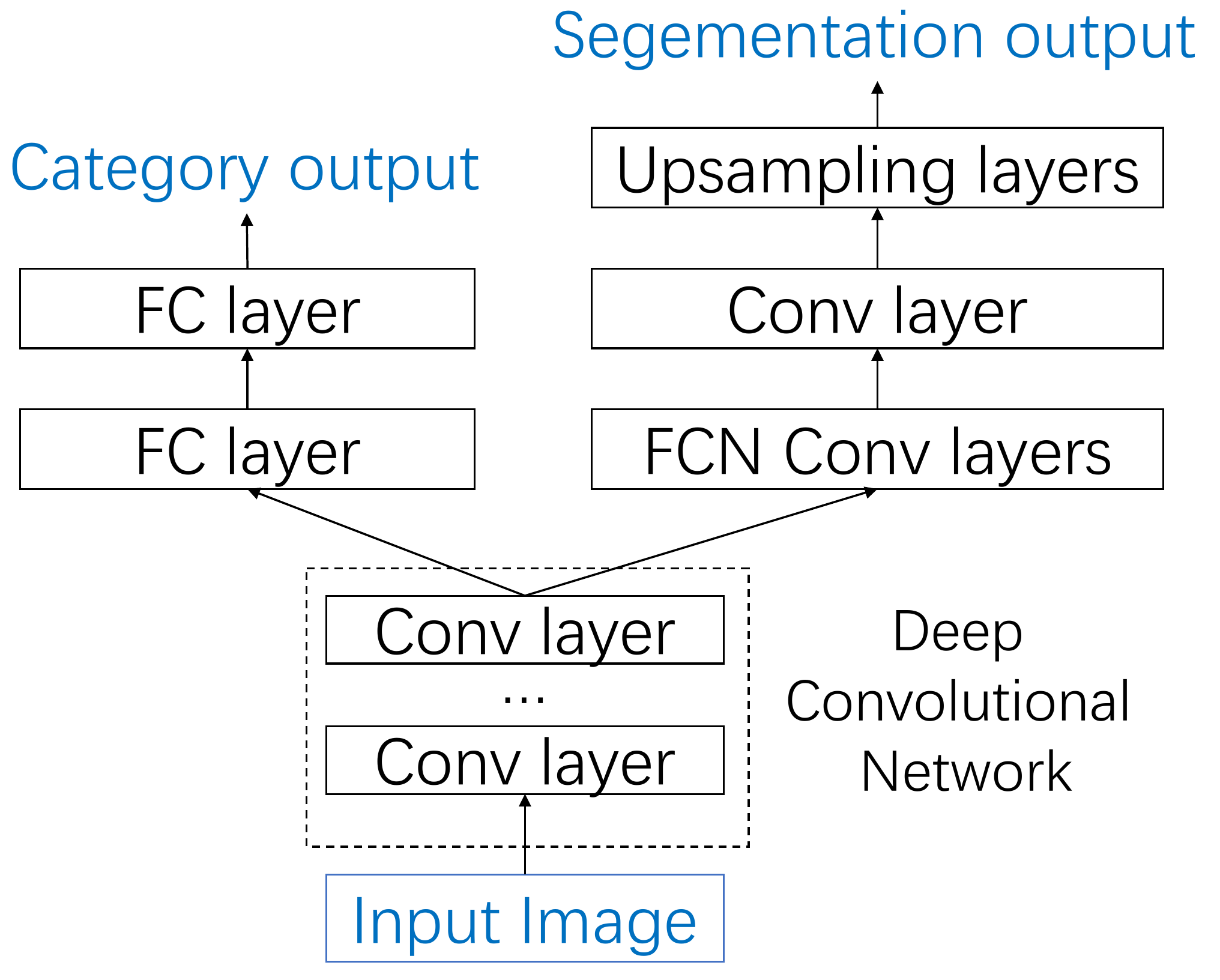}
      \label{fig:mfcn-a}
  }
  \subfigure[MFCN-B]{
      \includegraphics[scale=0.17]{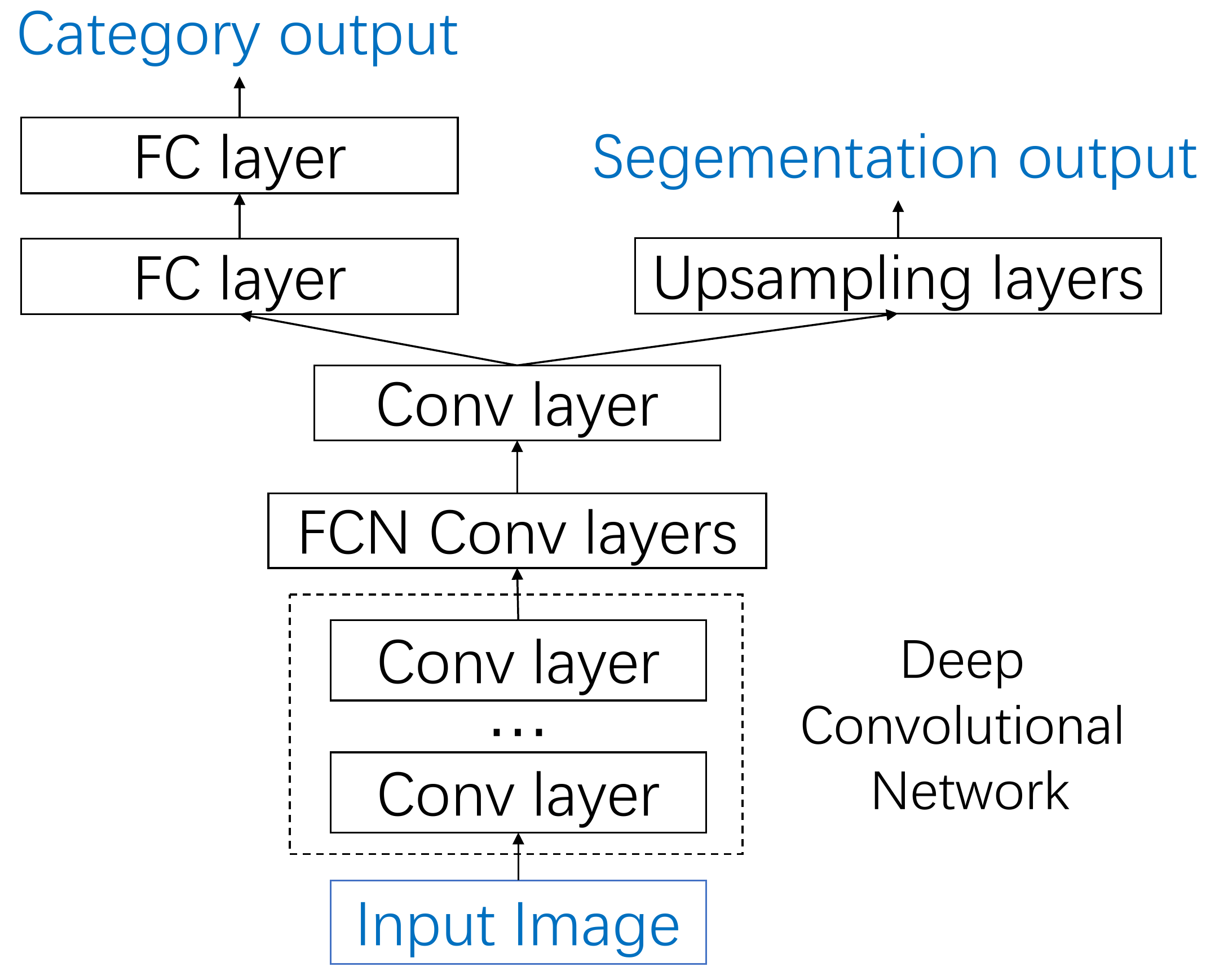}
      \label{fig:mfcn-b}
  }
  \subfigure[MFCN-C]{
      \includegraphics[scale=0.17]{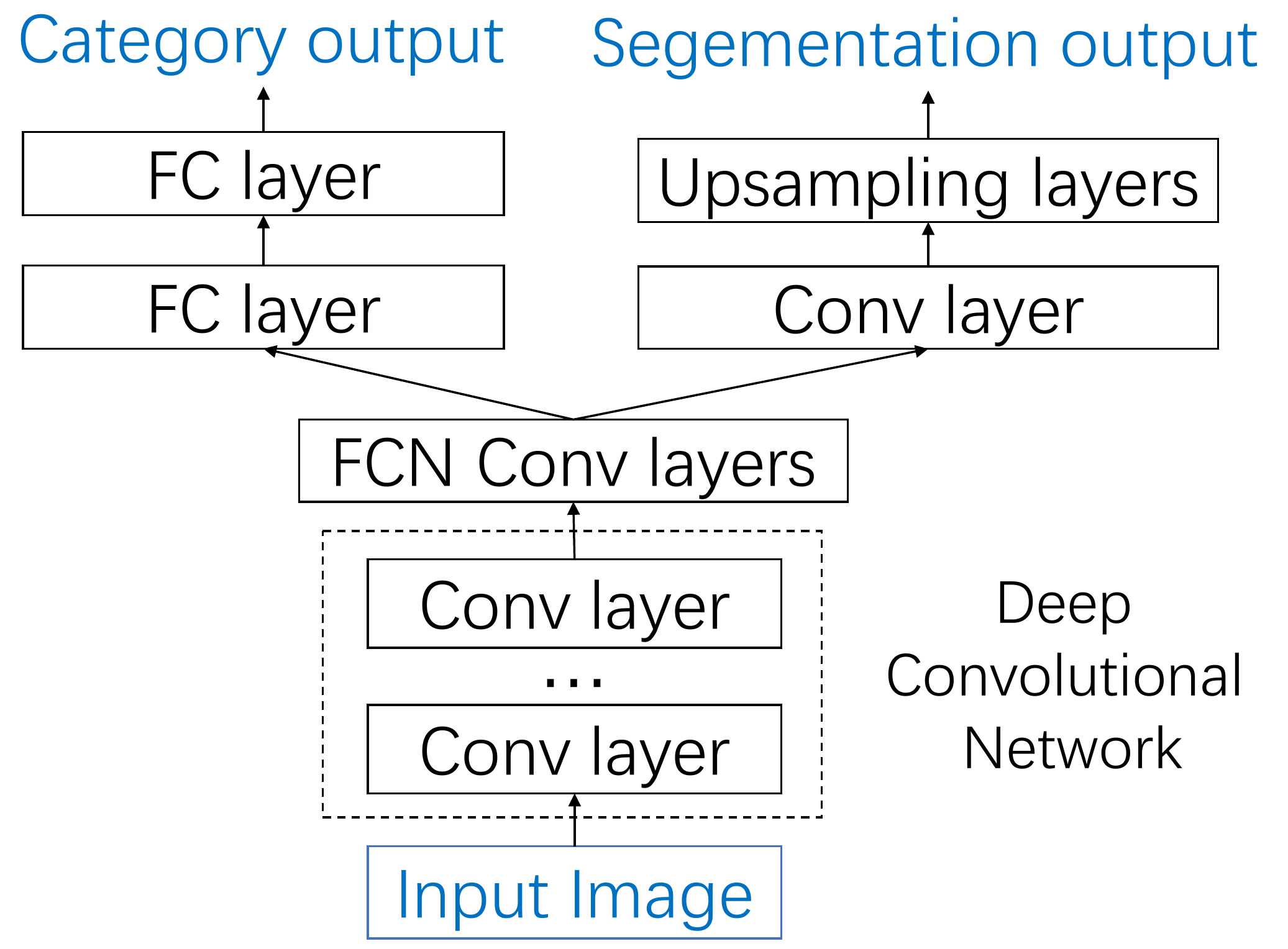}
      \label{fig:mfcn-c}
  }
  \caption{Three different deep architectures for multi-task learning of container classification and image segmentation}
  \label{fig:arch}
\end{figure*}

First, we use deep convolutional networks to extract convolution features of the image, and then input the extracted features into FCN for segmentation. Considering that the appearance of food in different containers may vary, sharing shape information of containers (e.g. bowls or plates) during the segmentation process may increase the accuracy of segmentation. To this end, we propose to combine the food segmentation task with the container classification task to jointly train the FCN.

We formulate food item segmentation and container classification as a multi-task deep learning problem and modify the architecture of the FCN for our purpose. Both tasks influence each other through updating the shared intermediate layers. The modification is not straightforward for involvement of a design issue, which is about the degree in which the intermediate layers should be shared. Ideally, each task should have its own private layer(s) given that the nature of both tasks, binary classification versus segmentation, is different. In such a way, the updating of parameters can be done more freely for optimization of individual performance.

Inspired by \cite{chen2016deep}, we derive three different deep architectures as depicted in Figure \ref{fig:arch}, respectively name as MFCN-A to MFCN-C.

\begin{itemize}
    \item MFCN-A allows classification network and FCN to privately own their intermediate layers on top of the convolutional layers for parameter learning.
    \item MFCN-B considers stacked architecture by adding the intermediate layers for container classification before upsampling the FCN extracted convolution features. In this setting, both tasks share all the convolution layers of FCN. Both MFCN-A and MFCN-B designs are relatively straightforward to implement by adding additional layers to FCN.
    \item MFCN-C considers the decoupling of some intermediate layers. It is a compromise version between the first and second architectures, by having shared convolution layers and private layers for each task.
\end{itemize}

We adopt two cross-entropy functions $L_{1},L_{2}$ as the loss functions for both tasks. Denote $N$ as the total number of training images, the overall loss function $L$ is as following:
\begin{equation}
    L=-\frac{1}{N}\sum_{n=1}^{N}{(L_{1}+\lambda L_{2})}
\end{equation}
where $\lambda$ is a hyper-parameter trading off the loss terms.

\subsection{Data Aggregation}\label{DataAggregation}
In this section, we build the food model and calculate food volume using segmented food items and the distance from the smartphone to the table obtained in the former section.

\subsubsection{Actual Size Scaling}
\begin{figure}[ht]
\centering
\includegraphics[scale=0.23]{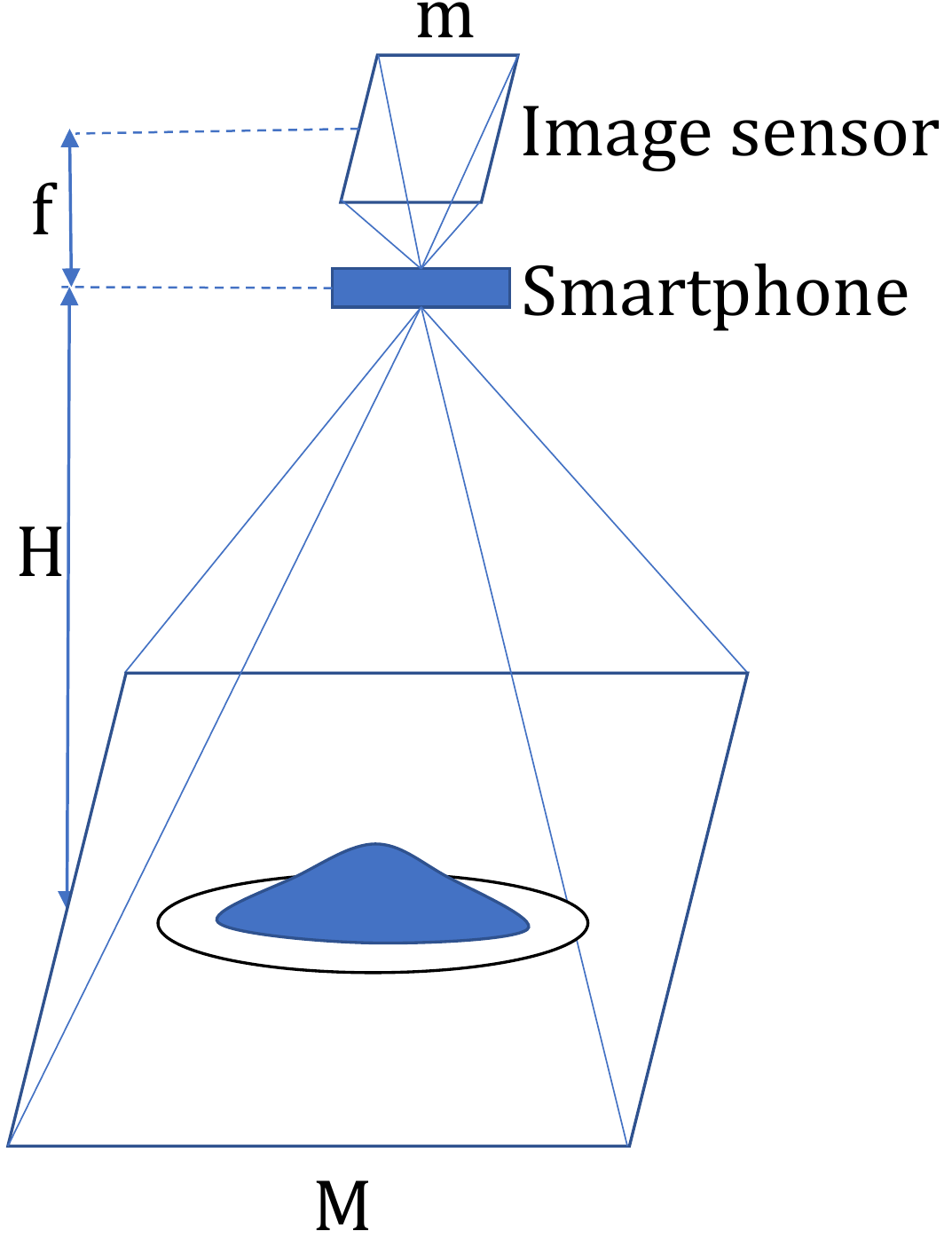}
\caption{Calculating the actual size of the image}
\label{fig:transmission principle}
\end{figure}

To get the actual size of the food in the image, we need to know the actual size corresponding to each pixel in the image. As shown in Figure {\ref{fig:transmission principle}}, according to the principle of scaling, the actual width of the image photoed at the vertical height $H$ can be calculated as: $M=\frac{{m}{H}}{{f}}$, where $H$ is the vertical distance from smartphone to table calculated in the former step, $f$ is the focal length of the smartphone lens, $m$ is the width of the smartphone image sensor(i.e. CCD or CMOS). For a certain smartphone, both $f$ and $m$ are fixed values.

Knowing the actual width of the image, we can easily get the actual size corresponding to each pixel in the image. Then the actual size of segmented food items in the top view image can be calculated by simply calculate how many pixels the food item has.

\subsubsection{Food Modeling And Volume Calculating}

\begin{figure}[ht]
\centering
\includegraphics[scale=0.13]{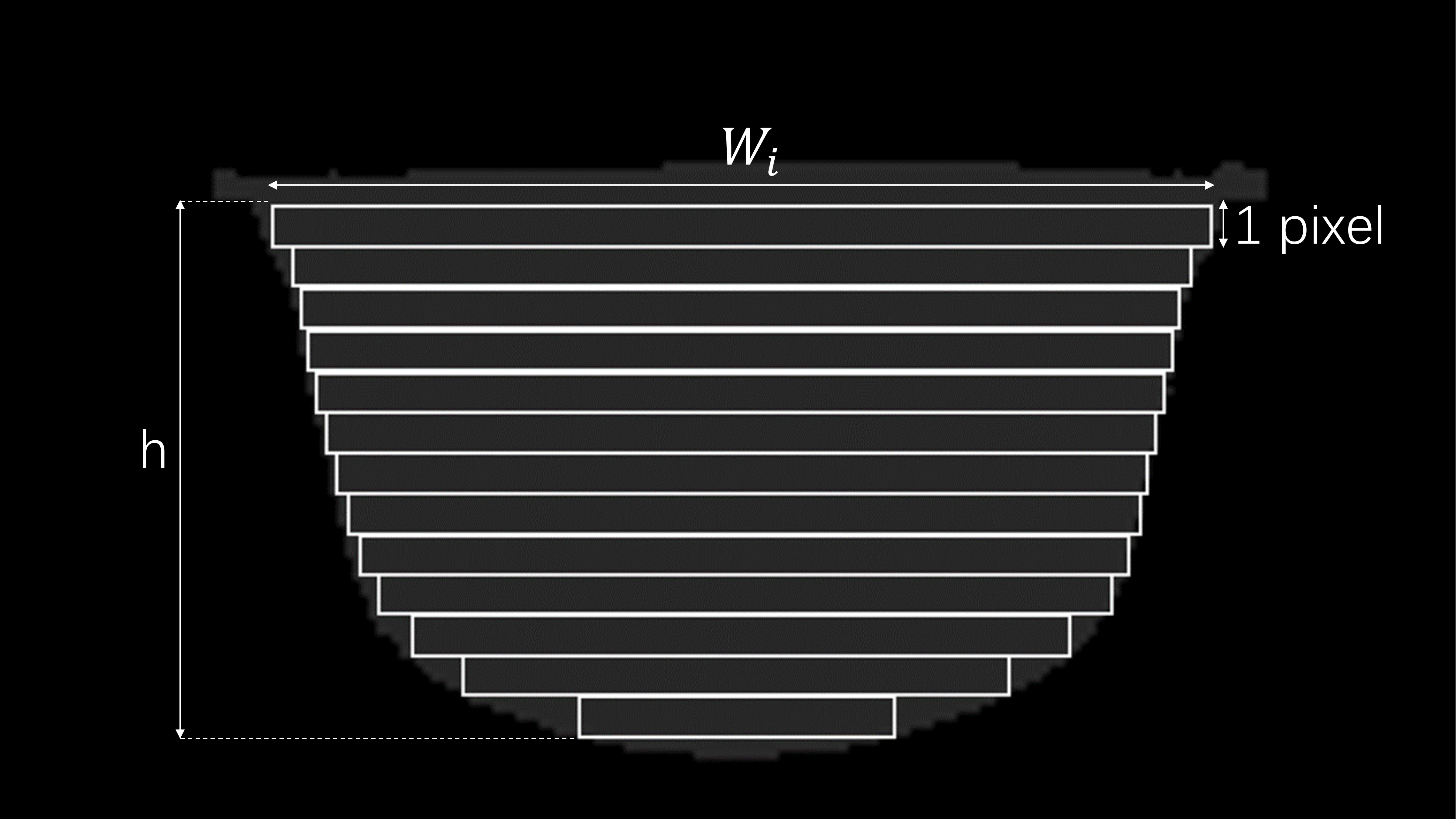}
\caption{Dividing food into columns with a height of 1 pixel}
\label{fig:volume_cal}
\end{figure}

After obtaining the actual food size in the top view image, we calculate food volume using the side view of the food based on a differential model. After segmenting the side image as well, We can obtain the side shape of the food (e.g. height and shape outlines) from the segmented food item. We assume that the food is uniformly distributed in the container (when the container is a bowl or a cup, the height of the food equals to the height of the container). As shown in Figure {\ref{fig:volume_cal}}, the food model can be regarded as the accumulation of columns with a height of 1 pixel, and the cross-sectional shape of each column is the shape of the food in the top view image. Assuming that the width of the segmented food item in the top view is $w_{f}$, the calculated actual size of food items in the top view is $S_{Real}$, the number of columns is $h$ (i.e. the height of the segmented food item in the side view image), and the width of the cross-section of the column $i$ is $w_{i}$.

The food volume $V$ can be calculated as:
\begin{equation}
    V=\sum_{i=1}^{h}{S_{Real}(\frac{w_{i}}{w_{f}}})^2
\end{equation}

\section{Evaluation}
We split the experiments into three parts. The first part aims to evaluate the accuracy of the MLS ranging method at different distances and the robustness under varying degrees of noise. The second part aims to evaluate different deep architectures of multi-task learning in comparison to single-task FCN and conventional CNN and \textit{Grabcut} method. The last part is designed to evaluate the final food volume estimation performance.

\subsection{MLS ranging}

The MLS we use has a length of 1023 samples (order $n=10$) and is generated using a sample rate of 48kHz. The 48kHz clock frequency is limited by the data acquisition hardware, and the length of the sequence is a compromise between resolution and the sequence period (approximately 20ms). The smartphone we use is an iPhone 6 Plus. Due to the hardware limitations of the phone speaker, playing with the bottom speaker can cause severe signal distortion, so the MLS signal must be played using the top speaker. We use the bottom microphone of the phone to record.

\begin{table*}[hbt]
\caption{The MLS ranging accuracy at different distances}
\begin{center}
\begin{tabular}{|c|c|c|c|c|c|}
\hline
&\multicolumn{5}{|c|}{\textbf{Distance}} \\
\cline{2-6} 
& \textbf{10cm}& \textbf{20cm}& \textbf{30cm}&\textbf{40cm}&\textbf{50cm} \\
\hline
\textbf{Measurement (cm)}&$10.08\pm0.00$ &$19.86\pm0.09$ &$29.74\pm0.12$ &$39.83\pm0.00$ &$49.68\pm0.08$ \\
\hline
\textbf{Relative error}&$(0.80\pm0.00)\%$ &$(0.70\pm0.49)\%$ &$(0.88\pm0.41)\%$ &$(0.42\pm0.00)\%$ &$(0.64\pm0.15)\%$ \\
\hline
\end{tabular}
\label{mls_acc}
\end{center}
\end{table*}

\subsubsection{Ranging accuracy}
In order to evaluate the accuracy of the MLS ranging method, we perform measurements between 10cm and 50cm. This is also the possible distance range between the phone and the desktop while the user is taking photos in actual use. We repeat the measurement 5 times at each distance. Table \ref{mls_acc} shows the results of our measurements using the MLS method at different distances. The results show that the MLS ranging method achieves high accuracy within the range of 10cm to 50cm. In addition, the distance itself has little effect on the accuracy of the MLS method.

\begin{table*}[hbt]
\caption{The longest retry time at different noise level}
\begin{center}
\begin{tabular}{|c|c|c|c|c|}
\hline
& \textbf{Dining room (23db)}& \textbf{Dining room (38db)}& \textbf{Restaurant (50db)}&\textbf{Cafeteria (68db)}\\
\hline
\textbf{The longest retry time (s)}&$0.0253\pm0.0038$ &$0.0265\pm0.0021$ &$0.0537\pm0.0343$ &$0.1048\pm0.0524$\\
\hline
\end{tabular}
\label{mls_noise}
\end{center}
\end{table*}

\subsubsection{Ranging robustness}

Environmental noise can have an impact on sound related methods. In order to evaluate the effect of different degrees of noise on the MLS method, we test the longest retry time of the MLS method ranging in environments of different level of noise. When the MLS signal is disturbed by noise, the ranging result value is usually far from the normal value. We repeat the MLS signal every 50ms until the correct distance is obtained, and we define the time spent in this process as the longest retry time. We test in four different situations, dining room (23db), dining room (38db), restaurant (50db) and cafeteria (68db). At each situation, we repeat the test 5 times.

Table \ref{mls_noise} shows the longest retry time of the MLS method at different noise level. The results show that although noise has a certain impact on the MLS method, this method still shows high robustness. In a quiet room, the MLS method can get an accurate result with just one measurement. In a relatively noisy environment, the MLS method may need to measure 2-3 times to get results. But the time spent on this process is very short and does not affect user experience. Furthermore, we find that in general situations, ranging results will not be affected by other objects on the desktop. However, if there are large objects on the table that are very close to the food, they may affect ranging results.

\subsection{Food image segmentation}
\subsubsection{Dataset}
We use our SUEC Food dataset as training images for food image segmentation. The images in SUEC Food are from the UEC Food-256 dataset\cite{kawano2014automatic}, which contains 256-kind food images with only food category labels and bounding boxes. Most of the food categories in UEC Food-256 are popular foods in Japan and other Asian countries. FCN requires training images with segmentation labels. In order to obtain the segmentation labels for UEC Food-256 images, we first segment the images in bounding boxes using the \textit{Grabcut} method. Then we manually segment 500 food images. We pre-train our models with \textit{Grabcut} generated segmentation images, then fine-tune the models with manually segmented images. In addition, we manually segmented another 100 food images as the test dataset. We have published this dataset, SUEC Food, including 31395 segmented images annotated by \textit{Grabcut} method and 600 segmented images annotated manually. For the container classification task, there are many types of containers. In this work, we divide the food containers into plates and bowls and label all the images manually.

\subsubsection{Experiment Setup}
Several state-of-the-art food volume estimation methods (e.g. \cite{liang2017deep, okamoto2016automatic}) use Fast R-CNN and \textit{Grabcut} to segment food images, which are taken as our baseline approaches. The \textit{Grabcut} method requires repeated iterations for each image to get the segmentation results. The more iterations, the longer it takes, and the more accurate the segmentation result. The iterations of \textit{Grabcut} are set to 1,2,3 and 5. In addition, in order to evaluate the multi-task models we proposed, we also use the original single-task FCN as one of our baselines.

We use VGG-16\cite{simonyan2014very} as the architecture of the deep convolutional neural networks. In the proposed MFCN, the hidden dimensions of the fully connect layers for container classification are set to 128 and 2. The convolutional layer before upsampling layers has 151 filters, and the kernel size is set to 1. The hyper-parameter $\lambda$ is set to 1. The learning rate is set to $1e-5$ with Adam optimizer.

\subsubsection{Evaluation Metric}
All the deep convolutional neural networks we use are pre-trained with 2000 categories in ImageNet\cite{deng2009imagenet}, including 1000 food-related categories. The networks are then trained with training images in SUEC Food. 90\% of the images are randomly picked for training, and 10\% for validation. We finally test all the models on the 100 test images with manually annotated ground truth labels.

Similar to other researches of image segmentation, we assess segmentation performance using the mean Intersection over Union ($mIoU$)
%: 
%\begin{equation}
%    mIoU=\frac{1}{k+1}\sum_{i=0}^{k}{\frac{p_{ii}}{\sum_{j=0}^{k}{p_{ij}}%+\sum_{j=0}^{k}{p_{ji}}-p_{ii}}}
%\end{equation}
%where $k+1$ is the number of classes (in our case, k=1), $p_{ij}$ is the amount of pixels of class $i$ inferred to belong to class $j$.

Besides, in order to evaluate the speed improvement brought by FCN, we segment 100 images using each method and record their time consuming.

\subsubsection{Performance}

\begin{table}[hbt]
\caption{Results of food image segmentation on test dataset}
\begin{center}
\begin{tabular}{|c|c|c|c|}
\hline
~&\textbf{Method}& \textbf{mIoU}& \textbf{Time (s)}\\
\hline
\multirow{5}*{\textbf{Baseline}}&\textbf{Grabcut@1}&$0.7428$ &$27.14$\\
\cline{2-4}
~&\textbf{Grabcut@2}&$0.7665$ &$36.56$\\
\cline{2-4}
~&\textbf{Grabcut@3}&$0.7720$ &$39.38$\\
\cline{2-4}
~&\textbf{Grabcut@5}&$0.7730$ &$54.27$\\
\cline{2-4}
~&\textbf{FCN}&$0.9143$ &$11.17$\\
\hline
\multirow{3}*{\textbf{Our approach}}&\textbf{MFCN-A}&$0.9160$ &\textbf{10.25}\\
\cline{2-4}
~&\textbf{MFCN-B}&\textbf{0.9210}&$12.62$\\
\cline{2-4}
~&\textbf{MFCN-C}&$0.9166$ &$13.30$\\
\hline
\end{tabular}
\label{segment}
\end{center}
\end{table}

\begin{table*}[hb]
\caption{Results of food volume estimation}
\begin{center}
\begin{tabular}{|c|c|c|c|c|c|c|}
\hline
\textbf{Container shape}&\multicolumn{4}{|c|}{\textbf{Plate}}
&\multicolumn{2}{|c|}{\textbf{Bowl}}\\
\hline
\textbf{Food}&\multicolumn{2}{|c|}{\textbf{Chicken drumstick}}
&\multicolumn{2}{|c|}{\textbf{Fried pork}}
&\multicolumn{2}{|c|}{\textbf{Congee}}\\
\hline
&\textbf{Estimation(ml)}&\textbf{Relative Error(\%)}&\textbf{Estimation(ml)}&\textbf{Relative Error(\%)}&\textbf{Estimation(ml)}&\textbf{Relative Error(\%)}
\\
\hline
\textbf{Real value}&$285.5$ &$0$ &$372.0$ &$0$ &$334.0$ &$0$\\
\hline
\textbf{Eye-measurement}&$246.0$ &$-13.84$ &$307.5$ &$-17.34$ &$410.4$ &$22.87$\\
\hline
\textbf{Prepaid card}&$151.7$ &$-46.85$ &$247.8$ &$-33.40$ &$213.8$ &$-35.98$\\
\hline
\textbf{Chopsticks(Known size)}&$313.5$ &$9.80$ &$446.9$ &$20.13$ &$356.1$ &$6.62$\\
\hline
\textbf{Chopsticks(Average size)}&$359.9$ &$26.04$ &$513.0$ &$37.90$ &$408.8$ &$22.40$\\
\hline
\textbf{Finger(Known size)}&$536.9$ &$88.05$ &$598.3$ &$60.83$ &$498.4$ &$49.22$\\
\hline
\textbf{Finger(Average size)}&$671.9$ &$135.06$ &$747.9$ &$101.04$ &$623.0$ &$86.52$\\

\hline\textbf{Our approach}&\textbf{293.2} &\textbf{2.70} &\textbf{418.0} &\textbf{12.37} &\textbf{333.1} &\textbf{-0.27}\\
\hline
\end{tabular}
\label{food_volume}
\end{center}
\end{table*}

Table \ref{segment} lists the results for food image segmentation. The $mIoU$ of FCN significantly outperformed all the \textit{Grabcut}-based methods. And the $mIoU$ of all three structures of MFCN are higher than the single-task FCN, which confirms our observation that the shapes of food contours are related with the shapes of containers. By sharing this information with food segmentation using multi-task learning, we can obtain a more accurate segmentation model. Specifically, among the three structures of MFCN, the MFCN-B structure performs best, followed by MFCN-C. The performance of MFCN-A is slightly lower than the other two structures, but still better than the single-task FCN. This is because there may be a close relationship between the shapes of food contours and the shapes of containers, and a large number of identical features are shared between the two tasks. Therefore, the more convolutional layers shared by the two tasks, the more accurate the final segmentation result. Besides, for the FCN and MFCN models, it takes only about 10 seconds to segment 100 images, while the fastest \textit{Grabcut} method (\textit{Grabcut}@1) takes 27 seconds. Different multi-task learning structures have no significant differences in time. The accuracy of the \textit{Grabcut} method is related to the time it takes. However, even if it takes more than 50 seconds, the accuracy of the \textit{Grabcut} method is only about 77\%. Thus, using FCN-based methods will significantly improve the efficiency of the whole food volume estimation process.

\subsection{Food Volume Estimation}

\begin{figure}[htb]
  \centering
  \subfigure[Chicken drumstick]{
      \includegraphics[scale=0.08]{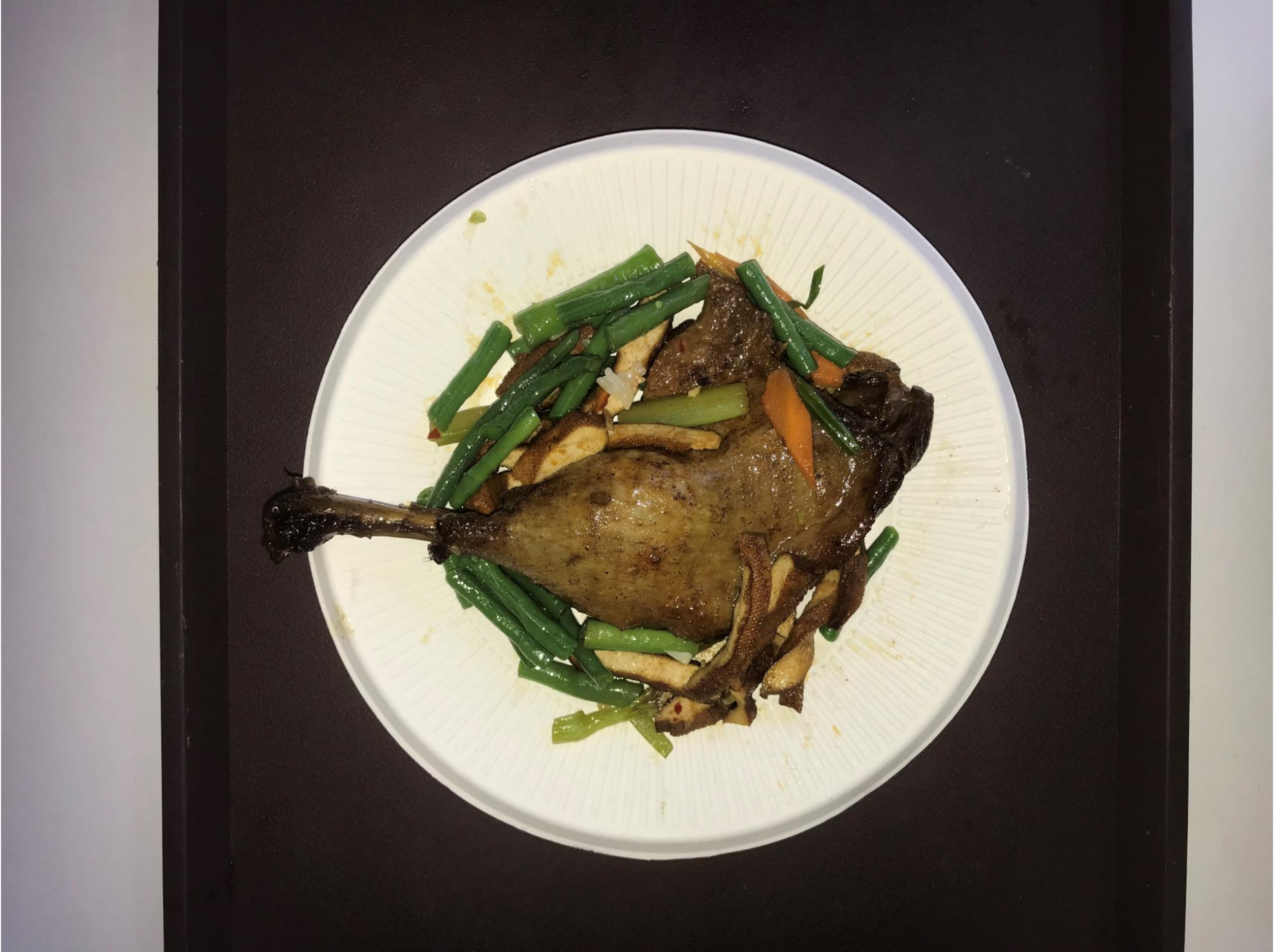}
      \label{fig:Chicken_drumstick}
  }
  \subfigure[Fried pork]{
      \includegraphics[scale=0.08]{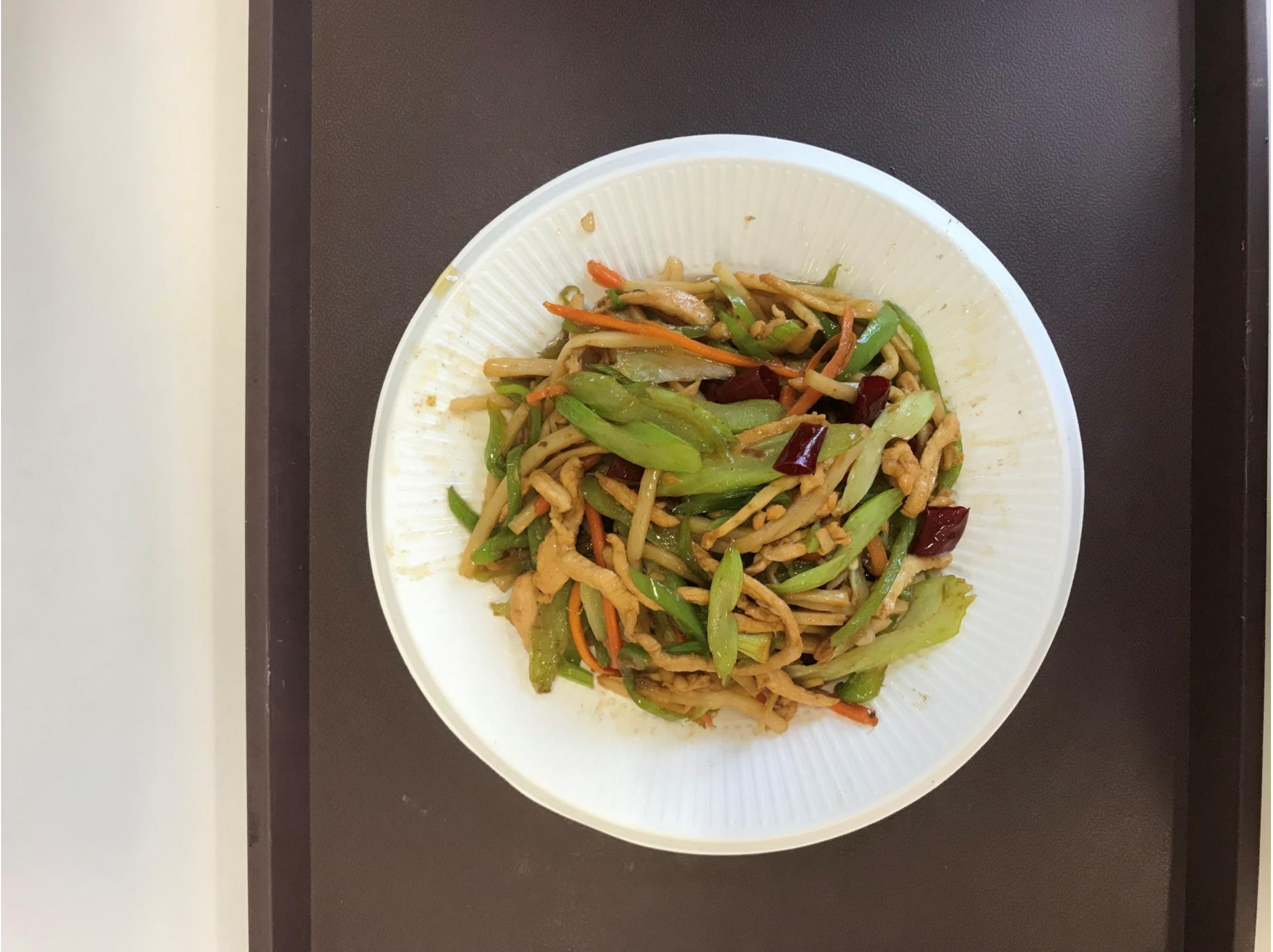}
      \label{fig:Fried_pork}
  }
  \subfigure[Congee]{
      \includegraphics[scale=0.08]{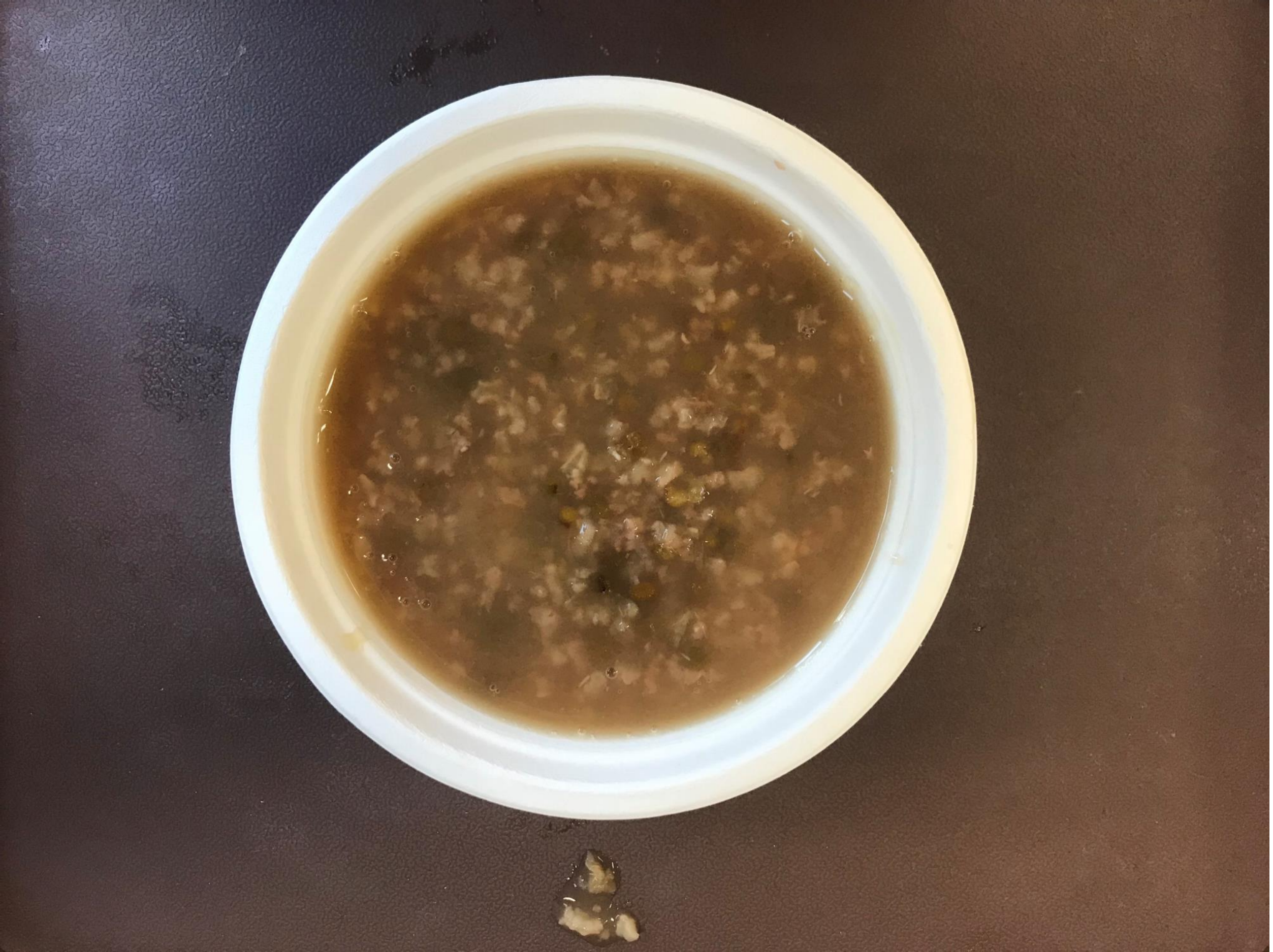}
      \label{fig:Congee}
  }
  \caption{Three foods for volume estimation testing}
  \label{fig:demo_dishes}
\end{figure}

We selected three foods for volume estimation, chicken drumstick(Fig. \ref{fig:Chicken_drumstick}), fried pork(Fig. \ref{fig:Fried_pork}) and congee(Fig. \ref{fig:Congee}), which are common home cooking. Chicken drumstick and fried pork are on the plates, and congee is in the bowl. The shapes of these foods vary, where chicken drumstick has a more regular shape and fried pork has a complex stacked shape. We use the water displacement method to get the actual volume of all foods.

We select three state-of-the-art reference objects-based food volume estimation models for comparison(i.e, Okamoto et al. 2016\cite{okamoto2016automatic}, Akpa et al. 2017\cite{akpa2017smartphoneF}, Pouladzadeh et al. 2016 \cite{pouladzadeh2016food}). For those methods which use chopsticks or fingers as reference objects, the sizes of chopsticks or fingers are usually very different. To be fair, we conduct two experiments using these methods. First, we use chopsticks or fingers of known length as the reference object and then use chopsticks or fingers of average length as the reference object (i.e. the average length of 10 different sizes of chopsticks or fingers). In addition, we also asked 10 people to visually estimate the food volume. We use the average of their estimation as one of our baselines. Table \ref{food_volume} lists the results of final food volume estimation.

The results show that \textit{MUSEFood} outperforms all the state-of-the-art methods for foods with different shapes in different containers. For foods with regular shapes and foods in bowls, the error of \textit{MUSEFood} is very low. For foods with irregular shapes, the error of \textit{MUSEFood} is slightly larger, but the accuracy is still far superior to all the baseline methods.

\section{Conclusion and Discussion}
In this paper, we have introduced \textit{MUSEFood} to calculate food volume using data collected from multiple sensors on smartphones. \textit{MUSEFood} uses FCN and utilizes shape information of food containers through multi-task learning structures, resulting in more accurate and fast food image segmentation. We use the MLS ranging instead of using reference objects, which improves the convenience of use and achieves higher food volume estimation accuracy.

\textit{MUSEFood} shows sufficient robustness and versatility in our experiments. The shape of the different food containers does not introduce errors in the results of the food volume estimation. \textit{MUSEFood} can handle food without regular shapes, which makes our models applicable to more types of foods. Though some smartphones have ToF cameras, which can directly obtain the target distance, smartphones with ToF cameras are not very popular. Requiring user to buy the special equipment will increase the cost inevitably. Furthermore, the develop interfaces of these ToF cameras are hardly public to developers. 
%For future development, we will further improve user experience, such as users can record a video instead of taking photos. It is easy to achieve by collecting the gyro sensor data of the phone while the user is recording the video, and select specific frames in the video according to the shooting angle of the user. This will further improve user friendliness. 
After obtaining the food volume, we can conveniently estimate the detailed calories and nutrients intake using food nutrients database. We will work on this function in the future.

\section*{Acknowledgment}
This work is supported by the National Science and Technology Major Project (No. 2018ZX10201002) and the National Natural Science Foundation of China (No.61772045). We gratefully acknowledge the support of NVIDIA Corporation with the donation of the Titan X Pascal GPU used for this research.

\bibliographystyle{IEEEtran}
\bibliography{IEEEabrv,mybib}

\end{document}